\documentclass{article}
\usepackage[english]{babel}
\usepackage[utf8]{inputenc}
\usepackage{johd}
\title{Evaluating the Evaluator: Problems with SemEval-2020 Task 1 for Lexical Semantic Change Detection}

\author{Bach Phan-Tat$^{a}$$^{*}$, Kris Heylen$^{a}$$^{b}$, Dirk Geeraerts,$^{a}$
        Stefano De Pascale,$^{a}$$^{c}$, Dirk Speelman$^{a}$ \\
        \small $^{a}$Department of Linguistics, KU Leuven, Leuven, Belgium \\
        \small $^{b}$Instituut voor de Nederlandse Taal, Leiden, The Netherlands \\
        \small $^{c}$Department of Linguistics and Literary Studies, Vrije Universiteit Brussel, Brussels, Belgium \\
        \small $^{*}$Corresponding author: Bach, Phan-Tat; \tt{ttbach.phan@kuleuven.be}
}

\date{} 

\begin{document}

\maketitle

\noindent{Discussion paper}

\begin{abstract}
\noindent This discussion paper re-examines SemEval-2020 Task 1, the most influential shared benchmark for lexical semantic change detection, through a three-part evaluative framework: \textit{operationalisation}, \textit{data quality}, and \textit{benchmark design}. First, at the level of operationalisation, the benchmark models semantic change mainly as gain, loss, or redistribution of discrete senses. We argue that while practical for annotation and evaluation, this framing is too narrow to capture gradual, constructional, collocational, and discourse-level change. Also, the gold labels are outcomes of annotation decisions, clustering procedures, and threshold settings, which could potentially limit the validity of the task. Second, at the level of data quality, we show that the benchmark is affected by substantial corpus and preprocessing problems, including OCR noise, malformed characters, truncated sentences, inconsistent lemmatisation, POS-tagging errors, and missed targets. These issues can distort model behaviour, complicate linguistic analysis, and reduce reproducibility. Third, at the level of benchmark design, we argue the small curated target sets and limited language coverage reduce realism and increase statistical uncertainty. Taken together, these limitations suggest that the benchmark should be treated as a useful but partial test bed rather than a definitive measure of progress. We therefore call for future datasets and shared tasks to adopt broader theories of semantic change, document preprocessing transparently, expand cross-linguistic coverage, and use more realistic evaluation settings. Such steps are necessary for more valid, interpretable, and generalisable progress in lexical semantic change detection.
\end{abstract}

\noindent\\

\noindent\authorroles{Bach Phan-Tat: Conceptualisation, Data curation, Formal analysis, Investigation, Methodology; 
\\
Kris Heylen: Formal analysis, Project administration;
\\
Dirk Geeraerts, Stefano De Pascale, Dirk Speelman: Conceptualisation, Methodology, Project administration
}\\

\section{Context and motivation}

\subsection{What is Lexical Semantic Change?}
Semantic change is a phenomenon of linguistics where word meanings shift over time. Meanings may become more specialised, as with \textit{queen}, which originally meant ‘wife’ or ‘woman’ but later narrowed to mean ‘the king’s wife’ or ‘a female sovereign.’ Meanings may also become more general, as with \textit{moon}, which once referred mainly to 'the Earth’s satellite' but was later extended to 'the satellites of other planets'. Semantic change can also occur through metonymy, where a word comes to stand for something closely associated with it, as in \textit{drink a whole bottle}. A diachronic example is English \textit{board}, which developed from the sense `table' to `the people sitting around a table' and then to `governing body' \citep{traugott_semantic_2017}. Likewise, metaphor can create new meanings through resemblance, as in \textit{grasp}, which developed from the physical sense of ‘hold’ to the abstract sense of ‘understand.’ In addition, words may shift in connotation/valence, becoming either more positive or more negative over time: \textit{nice}, for example, moved from ‘foolish’ to ‘pleasant,’ while \textit{boor} shifted from ‘peasant’ to ‘an unmannered person.’ (see \citet{geeraerts_theories_2010, geeraerts_semantic_2020} for an overview of Lexical semantics and classification of types of semantic changes). Semantic change is driven by many factors, including cultural change, technological innovation, social interaction, contact between languages, and speakers’ tendency to reuse old words for new situations \citep{traugott_semantic_2017}. This makes semantic change central to understanding language as a dynamic system rather than a static list of meanings.

\subsection{Lexical Semantic Change in Computational Linguistics/Natural Language Processing}
With the development of historical text corpora, computational and statistical techniques, and the recent advancement of neural networks, linguists have more resources for investigating semantic change on an automatic and/or large-scale manner. In Computational Linguistics and Natural Language Processing, lexical semantic change is usually operationalised as a measurable difference in a word’s usage across time-sliced corpora. Rather than observing meaning change directly, researchers model it through changes in distribution: the differences in the different linguistic contexts between 2 periods are taken as evidence of semantic changes \citep{tahmasebi_computational_2021}. This development has also created the need for shared tasks and benchmark datasets, which provide a common basis for evaluating systems in a more comparable and systematic way, rather than relying only on isolated case studies \citep{schlechtweg_semeval-2020_2020}.

To create these benchmark datasets, human annotation is needed. A particularly influential annotation framework in this area is DURel, or Diachronic Usage Relatedness \citep{schlechtweg_diachronic_2018}. Instead of asking annotators to assign predefined dictionary senses to word usages, DURel asks them to judge the semantic relatedness between pairs of usages of the same target word, usually on an ordinal scale. These pairwise relatedness judgments can then be used to infer whether the meaning of a word has remained stable or changed across time. For example, if usages from an earlier period and usages from a later period are consistently judged as less related to each other than usages within the same period, this provides evidence for lexical semantic change.

\subsection{Lexical Semantic Change Benchmarks}
Some popular shared tasks and datasets in lexical semantic change are SemEval 2020 Task 1 \citep{schlechtweg_semeval-2020_2020}, DIACR-Ita \citep{basile_diacr-ita_2020}, RuShiftEval \citep{kutuzov_grammatical_2021}, LSCDiscovery \citep{zamora-reina_lscdiscovery_2022}, TempoWiC \citep{loureiro_tempowic_2022}, NorDiaChange \citep{kutuzov_nordiachange_2022}, ChiWUG \citep{chen_chiwug_2023}, AXOLOTL’24 \citep{fedorova_axolotl24_2024}. Most of them use the DURel framework \citep{schlechtweg_diachronic_2018} for human annotations when creating the dataset. Amongst them, SemEval 2020 Task 1 is arguably the most influential benchmark, since it introduced the first shared evaluation framework for the task, provided high-quality multilingual gold data, and attracted 33 teams submitting 186 systems. Since its introduction, it has been employed in the evaluation of many different systems \citep{laicher_explaining_2021, kutuzov_grammatical_2021, kutuzov_contextualized_2022, cassotti_xl-lexeme_2023, zhou_finer_2023, ma_graph-based_2024, oda_improving_2025, phan-tat_reframe_2026, phan-tat_transparent_2026}. 

In this paper, we therefore critically revisit SemEval-2020 Task 1 through three analytical lenses: operationalisation, data quality, and benchmark design, to examine how far it constitutes a reliable benchmark for lexical semantic change detection.

\section{Dataset description}
Below is a quick description of the dataset and its creation process, which is discussed in detail in their original paper \citep{schlechtweg_semeval-2020_2020}, and their metadata.

SemEval 2020 Task 1 covers four languages: English, German, Swedish, and Latin. Each language is associated with its own curated corpus and set of target lexemes. For English, the dataset uses the Clean Corpus of Historical American English (CCOHA) \citep{alatrash_ccoha_2020, davies_expanding_2012}, which spans 1810s–2000s. For German, it draws on the DTA corpus \citep{deutsches_textarchiv_grundlage_2017}, together with the BZ and ND corpora \citep{staatsbibliothek_zu_berlin_berliner_2018, staatsbibliothek_zu_berlin_neues_2018}. DTA includes texts from a range of genres covering the 16th to the 20th centuries, while BZ and ND are newspaper corpora spanning the period 1945–1993. For Latin, the dataset uses the LatinISE corpus \citep{mcgillivray_tools_2013} which ranges from the 2nd century BC to the 21st century AD. For Swedish, Kubhist corpus \citep{sprakbanken_text_kubhist_2019}, a newspaper corpus covering the 18th to the 20th centuries was used. All corpora are lemmatized and POS-tagged. In addition, CCOHA and DTA are spelling-normalized, whereas BZ, ND, and Kubhist contain frequent OCR errors.

All datasets were sampled and separated into 2 sub-corpora, C1 and C2. The organisers then selected some target lexemes: 37 words for English, 48 for German, 40 for Latin and 31 for Swedish; and annotate them using the DURel framework.

Due to licensing restrictions, we do not redistribute the full corrected version of the dataset. Instead, we provide links to:
\begin{itemize}
    \item the original dataset
    \item the code and correction procedure used to reproduce the pre-processing steps in our previous study \footnote{\url{https://github.com/phantatbach/LChange26-Dep}}
    \item the sample, along with the code and annotation used in the qualitative analysis in this study \footnote{\url{https://doi.org/10.5281/zenodo.19408459}}
\end{itemize}

\subsection{Original dataset}
\paragraph{Repository location}
\url{https://www.ims.uni-stuttgart.de/en/research/resources/corpora/sem-eval-ulscd/}
\paragraph{Repository name}
Test Data for SemEval-2020 Task 1: Unsupervised Lexical Semantic Change Detection
\paragraph{Object name} \begin{itemize}
    \item English: semeval2020\_ulscd\_eng.zip
    \item German: semeval2020\_ulscd\_ger.zip
    \item Latin: semeval2020\_ulscd\_lat.zip
    \item Swedish: semeval2020\_ulscd\_swe.zip
\end{itemize}
\paragraph{Format names and versions} CSV, TXT.
\paragraph{Creation dates} 2020
\paragraph{Dataset creators} Dominik Schlechtweg, Barbara McGillivray, Simon Hengchen,
Haim Dubossarsky, Nina Tahmasebi.
\paragraph{Language} English, German, Swedish, Latin
\paragraph{License} CC BY 4.0 (Latin, Swedish)
\paragraph{Publication date} 2020

\section{Discussion}

\subsection{Lexical Semantic Change operationalisation}

\subsubsection{Lexical Semantic Change or sense inventory change?}
Many lexical semantic change benchmarks operationalise change through a reduced analytical framework which treats lexical semantic change as change in sense inventory in order to make annotation and evaluation feasible. SemEval-2020 Task 1 is an influential example of this broader design logic. In subtask 1, a lexeme is assigned the label of 1 if it has gained or lost a sense. In subtask 2, the degree of change of a lexeme is determined by the divergence of its sense distributions between 2 periods \citep{schlechtweg_semeval-2020_2020}. This formulation has clear practical advantages: it provides annotatable units, produces explicit gold labels, and allows systems to be compared quantitatively. However, it also narrows the phenomenon of lexical semantic change to one particular representation of meaning, namely discrete sense inventories. The issues discussed below are therefore not unique to SemEval 2020, but reflect broader methodological trade-offs in lexical semantic change detection benchmarking.

The first limitation concerns the discreteness of senses. Sense boundaries are rarely sharp \citep{cruse_lexical_1986, geeraerts_vaguenesss_1993, kilgarriff_i_1997}. Many usages lie between established senses, and diachronic change often proceeds gradually through contextual extension, pragmatic strengthening, or shifts in prototype structure \citep{geeraerts_theories_2010, hanks_lexical_2013, traugott_semantic_2017}. If meaning changes continuously, then a before/after comparison of sense inventories imposes an artificial categorical structure on a gradient process.

The second limitation is methodological. A sense inventory is not directly observable in the corpus but an analytical construct inferred from usage data. In the case of SemEval, annotators first judge the semantic relatedness between usage pairs (DURel framework), and these judgements are then transformed into usage graphs, clusters, sense assignments, and finally gold scores. This means that the resulting gold labels reflect not only diachronic usage patterns, but also annotation uncertainty, clustering procedures, and thresholding decisions. The moderate inter-annotator agreements reported in SemEval 2020 illustrates this point. The official report gives overall agreement of roughly 0.52 for English, 0.60 for German, and 0.58 for Swedish, while Latin posed additional difficulties because annotators often had to translate excerpts and each Latin target was assigned to only one annotator due to the scarcity of qualified annotators \citep{schlechtweg_semeval-2020_2020}. In our previous study, we have found one overlooked change by the organisers, which is the German noun \textit{Seminar}. Its official gold label for Subtask 1 is 0 (i.e., no sense gain or loss) and its official Subtask 2 rank is only 39/48 (i.e., low degree of semantic change) but we observe strong evidence for a metonymic shift from \textsc{institutional department} to \textsc{courses} \citep{phan-tat_transparent_2026}. Note that these do not invalidate the benchmark. Rather, they show that 1) sense assignment is a difficult interpretive task, especially in diachronic settings where usage contexts may be historically distant, noisy, or unfamiliar; and 2) benchmark results should be interpreted with awareness that the gold labels reflect a particular analytical procedure and may underrepresent certain kinds of change.

A third limitation is that many sense-inventory benchmarks do not make explicit the usage-level evidence on which a semantic-change diagnosis is based. Such evidence may include shifts in collocates, syntactic environments, constructional patterns, registers, or discourse domains. SemEval 2020 partly incorporates usage variation through DURel-style annotation, but the final benchmark representation is still reduced to lemma-level sense clusters, binary change labels, or graded sense-distribution divergence. As a result, the benchmark tells us whether a lexeme is judged to have changed, but it does not systematically expose which contextual, collocational, syntactic, or discourse-level patterns support that judgement. This makes evaluation less explanatory: systems are rewarded for matching the final change label or ranking, but not for identifying the linguistic evidence that motivates the label. A richer benchmark could therefore retain sense change as the primary target while adding complementary explanatory subtasks that ask models to identify the usage-level evidence associated with the detected change.

Finally, even within a sense-inventory view of semantic change, some benchmarks leave important distinctions unresolved. For example, SemEval 2020 Subtask 1 asks whether a word has gained or lost a sense, but does not distinguish gain from loss as separate types of change. Later resources such as LSCDiscovery address this issue more directly by focusing on the discovery of lexical semantic change and allowing finer-grained distinctions in change type \citep{zamora-reina_lscdiscovery_2022}.

\subsubsection{Temporal restriction}
A further limitation is the time-restricted nature of the task. By evaluating semantic change only between 2 periods, the benchmark reduces a potentially continuous and multi-stage diachronic process to a single pairwise comparison. This design is practical for annotation and evaluation, but it may obscure when a change begins, whether it proceeds gradually or abruptly, and whether intermediate stages or reversals are involved. As a result, the task captures net difference between two periods rather than the full trajectory of semantic development.

\subsection{Data quality}

\subsubsection{Corpus Noise and Preprocessing Issues}

We first describe a range of problems that we identified while working with the data, and then discuss their potential negative effects on systems that are trained, evaluated, or tested on this dataset.

SemEval 2020 Task 1 does not apply a consistent denoising strategy across languages. While CCOHA, DTA and LatinISE were normalised or inherited prior cleaning efforts, BZ, ND, Kubhist were released with substantial residual noise. 

The English data come from CCOHA, which is itself a cleaned version of COHA created to address problems such as inconsistent lemmas and malformed tokens. That cleaning happened at the source-corpus level, not as a special SemEval denoising pass. Upon doing qualitative check with the English dataset, we found several problems: namely odd characters (e.g., \texttt{/z/, //}), OCR noises (e.g., \texttt{Id /z/ atiireclwiih giiuiunk down , a.s if to hidQ Ihcmselres ftom obaeivQlion ; tut all was perfectly still .}), file tags of the original CCOHA file (e.g., \texttt{@@555586 txt …}), odd sentence cut off (e.g., \texttt{of saying * of a calyx , that it is deeply cleft , the most proper language * obviously is … but one lateral sepal .}). Note that the unusual sentence cut-offs are partly caused by CCOHA's copyright-related replacement procedure. For copyright reasons, some tokens in CCOHA are replaced with a sequence of ten @ symbols. As an initial preprocessing step, the SemEval organisers split sentences around these replacement tokens and then removed the tokens, which sometimes resulted in unnatural sentence boundaries.

Another problem, which is a major one for English, is the erroneous POS tagging and lemmatisation by the organisers. In English, many target words show POS-specific semantic change so specifically in the English corpus (and not in the other 3 corpora), the target words were combined with broad POS tags (in the form target\_pos). However, many of the instances of the target lemmas appear to be incorrectly tagged, where VERBs were tagged as NOUNs and vice versa. For instance, in \texttt{… they sail round an island without land\_nn …}, \textit{land} is marked as a noun and appears as the target form \textit{land\_nn} even though in the original sentence it is actually a verb: \texttt{… They sail round an Island without landing. …}; or \textit{face} was tagged as a noun in \texttt{… and the face\_nn and backing …} (original context: \texttt{… and the facing and backing …}); or \textit{stroke} was tagged as a verb when it is actually a noun in \texttt{… a bell which strike four stroke\_vb …} (original context: \texttt{… a bell , which struck four strokes …}). Amongst the English target words, the target noun \textit{land} is the most problematic one, with many instances of such error. We occasionally found some instances where the organisers also missed the target words. For example, \textit{word} was not tagged with the suffix \textit{\_nn} in \texttt{… with the warning word her aunt …} (original sentence: \texttt{… with the warning words her aunt …}). 

For German, we found similar problems as that of the English corpora, such as odd characters (e.g., \texttt{Allü « rt ""} or \texttt{? \_•> 19}), clipped sentences (e.g., \texttt{der KPD 1935 und 1939 in den genannten Städten .}) or even incorrectly merged sentences from newspaper columns (e.g., \texttt{Abenteuerfilm aus 20.00 Fußball-WM 1986 18.30 An den Wasserfällen ...}) and OCR noises (e.g., \texttt{Onginalversuch} instead of \texttt{Originalversuch}; \texttt{Pro2ent} instead of \texttt{Prozent}). The original paper says the later-period (GER 2) newspaper data (BZ + ND) still contained frequent OCR noise. There is no evidence in the task paper showing dedicated OCR-correction stage to BZ/ND data. We also found 59 empty lines in the lemmatised files (36 in the first period and 23 in the second), which does not happen with the other 3 languages. Interestingly, in GER 1, we found 2 French sentences: \texttt{Mais si quelque ... de la Russie .} and \texttt{Recberches experimentales ... av. planches .}. This issue of off-target-language sentences is not observed in the other three languages.

For Swedish, the released task data appear to contain substantial residual OCR noise. The Swedish dataset page explicitly says KubHist contains very frequent OCR errors, especially in older data (SWE 1) and words without a lemma were simply left as they were \citep{sprakbanken_text_semeval2020_2024}. These OCR-related issues are prominent in the data files, for example in forms such as \texttt{bloqiiade-cillstZnd} instead of \texttt{blockade-tillstånd}. Other prominent problems, which could also be lumped into OCR-related errors, include anomalous characters (e.g. \texttt{\&gt;}), corrupted word boundaries (e.g. \texttt{JnteckningShaswareS}), and truncated sentence fragments (e.g. \texttt{hwilka klander häremot i laga ti » «n-}).

For Latin, the LatinISE dataset page says they were partially corrected by hand \citep{mcgillivray_latinise_2020}. Perhaps that is why our qualitative investigations showed no prominent problems like that of German or Swedish data.

To provide a quantitative indication of these problems, we analysed the first 100 rows from each period for each language. Since the released data had already been randomly shuffled, these samples are unlikely to reflect positional bias. We then used Gemini 2.5 Flash using the official Google paid API \footnote{Our aim was only to provide a \textbf{simple quantitative indication} of the data quality issues that we had already identified qualitatively. The categories considered here are mostly surface-level problems, such as sentence cut-offs, sentence merges, OCR noise, abnormal characters, broken word boundaries, and metadata remnants, rather than fine-grained linguistic or semantic annotations. We therefore used Gemini as a first-pass detector to make the procedure more scalable.} to detect the relevant issues, then manually verifying before counting, as shown in table \ref{tab:data_issues}. The results indicate that abnormal sentence boundaries, including cut-off and oddly merged sentences, are pervasive across ENG 1 and 2 (CCOHA), GER 2 (BZ + ND) AND SWE 1 (Kubhist). OCR noise and unusual characters are also frequent in SWE 1 and 2 (Kubhist) and GER 2 (BZ + ND). Latin, by contrast, appears to be the cleanest subset, with the fewest detected errors. These results are consistent with the qualitative observations above and with the different levels of source-corpus cleaning reported for the original datasets.

\begin{table}[ht]
\centering
\begin{tabular}{lrrrrrr}
\hline
\textbf{File} & \textbf{S Cutoff} & \textbf{S Merged} & \textbf{T OCR} & \textbf{T Boundary} & \textbf{T Chars} & \textbf{Metadata} \\
\hline
ENG 1 & 19 & 0  & 7   & 4  & 9  & 1 \\
ENG 2 & 10 & 2  & 0   & 9  & 4  & 0 \\
GER 1 & 3  & 0  & 2   & 0  & 2  & 0 \\
GER 2 & 13 & 5 & 68  & 21 & 11 & 2 \\
LAT 1 & 3  & 1  & 0   & 0  & 0  & 5 \\
LAT 2 & 5  & 1  & 1   & 2  & 3  & 8 \\
SWE 1 & 37 & 0  & 212 & 23 & 40 & 0 \\
SWE 2 & 9  & 4  & 90 & 6 & 7  & 0 \\
\hline
\end{tabular}
\caption{Counts of data quality issues by language-period subset. S = Sentence, T = Token}
\label{tab:data_issues}
\end{table}

Odd characters, OCR noise, and sentence cut-offs can distort Semantic Change Detection systems. Odd characters create false word forms and fragment frequency counts, so the model may treat the same word as several unrelated items. OCR noise introduces spelling errors that corrupt context windows, embeddings, and collocation patterns, making semantic shifts look stronger or weaker than they are. Sentence cut-offs remove or truncate the surrounding context, which is especially harmful for context-based models that infer meaning from nearby words. The wrong lemmatisation and POS in the case of English, along with the missed targets also raise concerns about the \textbf{reliability of the gold score}, as they were computed using sentences containing the erroneous target\_POS pairs.

\subsubsection{Data reparsing}
Systems relying not just on co-occurrence information, but more on linguistic annotations \citep[e.g.,][]{kutuzov_grammatical_2021, phan-tat_transparent_2026} would require the dataset to be reparsed. Yet this is not a straightforward solution. The first caveat is spelling variation. Historical corpora often contain substantial orthographic variation, and SemEval 2020 data inherit many of these inconsistencies from their source corpora. As a result, different parsers especially those trained on different historical treebanks, may normalise or analyse the same form differently, assigning different lemmas to orthographic variants of the same word. Even though SemEval organisers claim that CCOHA and DTA were spelling-normalised \citep{schlechtweg_semeval-2020_2020}, they never released their normalisation procedure. This makes it difficult to align lexical items consistently across periods and, in turn, to measure semantic change reliably. A notable example is the German noun \textit{Lyzeum}, which was consistently lemmatised as \textit{Lyceum} (which was the old spelling) in the first period. Another example is the German verb \textit{artikulieren}, which has different surface forms in the raw corpora and as a result was lemmatised differently by Stanza or UDPipe (e.g., \textit{artikulieren , articulerien, articulieren, artikuliren, artikulir, artikulienen, articulirt}). This explains why \cite{kutuzov_grammatical_2021} only found 3 instances of the verb in the 19th century. Another problem we found when reparsing the data was the POS differences. Many instances of the target nouns in SemEval were tagged as proper nouns by Stanza and many target verbs were tagged as adjectives because they function as participles (e.g., \textit{antydd, andtydand} for the Swedish verb \textit{antyda}). 

The impact of these issues has been quantified in our previous case study, where we constructed two variants of the dataset, SemEval-faithful and Stanza-faithful, and compared the results obtained from each. The SemEval-faithful variant preserves the original SemEval target lemma and POS information as much as possible, correcting the parser output when it conflicts with the official target annotation. By contrast, the Stanza-faithful variant follows Stanza's own lemmatisation and POS-tagging decisions, with only minimal manual correction. Comparing these two variants allowed us to estimate how much performance depends on whether one follows the benchmark annotation or the output of a modern parser. When using the SemEval-faithful variant of the dataset, all of our models consistently achieved higher performance \citep{phan-tat_transparent_2026}. To our knowledge, our previous study is the only one to examine these issues explicitly and in detail across all four languages, as well as to document the corresponding pre-processing procedures. Based on our findings, we argued that future benchmarks and systems should treat pre-processing as a crucial step and report their procedures explicitly to ensure transparency and reproducibility.

We also acknowledge that for historical linguistic analysis, orthographic variation may itself be meaningful, since different spellings can reflect period-specific conventions, standardisation processes, genre differences, or source-specific practices. If only the normalised form is reported, researchers may mistakenly treat it as the form actually attested in the historical corpus. However, for the specific task of Lexical Semantic Change Detection, some degree of normalisation is often necessary: without grouping orthographic variants of the same lexeme, frequency counts become fragmented and semantic change estimates become less reliable. The crucial point is therefore not that normalisation should be avoided, but that both raw and normalised forms, together with the normalisation procedure, should be \textbf{documented explicitly}.

\subsection{Benchmark design}

\subsubsection{How many are enough?}
The first issue is benchmark realism. The task does not ask systems to find changing words in the full vocabulary. Instead, it gives a small pre-selected target list of words per language, which were chosen by scanning historical dictionaries for hypothesized changes, then pairing them with manually checked for POS and frequency development. The released test sets are correspondingly small: 37 targets for English, 48 for German, 40 for Latin, 31 for Swedish. Later work explicitly argues that performance on such small curated sets does not directly test whether a system can discover semantic change in realistic large-vocabulary settings \citep{umarova_current_2025, zamora-reina_lscdiscovery_2022}.

The next issue lies in the statistical nature of the small test set. These small test set makes the evaluation process statistically coarse and potentially noisy. Small differences in reported accuracy or correlation may reflect sampling noise rather than robust differences in system qualities.

In subtask 1, a correct identification of a single target, as either true negative or positive, changes the accuracy by 2.1--3.2 percentage points (i.e., \(1/n\), where \(n\) is the number of target lemmas). This makes performance differences look larger and more meaningful than they really are. For example, the difference between \(20/37\) (\(0.54\)) and \(24/37\) (\(0.649\)) appears substantial, even though it corresponds to only four lexemes. 

We could further illustrate this problem using the standard margin of error (MoE). For subtask1, the MoE formula would be:

\[
\mathrm{MoE} = 1.96 \cdot \sqrt{\frac{p(1-p)}{n}}.
\]

Here, \(p\) is the observed accuracy, \(n\) is the number of evaluated target words, and \(1.96\) is the critical value corresponding approximately to a 95\% confidence interval under a normal approximation. The term \(p(1-p)\) represents the variance of a binary correct/incorrect decision. This variance is largest when \(p = 0.5\), which is why using \(p = 0.5\) gives a worst-case or upper-bound estimate of the margin of error.

Under this approximation, the worst-case 95\% margins of error for the four SemEval 2020 languages are large, ranging from roughly 14.1 to 17.6 percentage points. For English, where \(n = 37\), the margin of error is approximately

\[
1.96 \cdot \sqrt{\frac{0.5(1-0.5)}{37}} \approx 0.161.
\]

This means that an observed accuracy of 50\% would correspond to an approximate interval of 33.9\% to 66.1\%. This means that a model that appears slightly above or below another model may not be meaningfully different in true performance.

A similar issue arises in Subtask 2, where systems are evaluated by Spearman's rank correlation between predicted and gold change rankings. Unlike accuracy, which can be treated as a binomial distribution, the sampling uncertainty of a correlation coefficient is usually approximated after transforming the correlation to a scale on which standard errors are easier to compute. A common approximation is Fisher's \(z\)-transformation:

\[
z = \frac{1}{2} \ln\left(\frac{1+r}{1-r}\right).
\]

The Fisher transformation creates a new statistic (\(z\)) whose sampling distribution is approximately normal under standard assumptions, making it easier to approximate standard errors. This makes it easier to approximate the standard errors. The standard error of the transformed value is approximately

\[
\mathrm{SE}_z = \frac{1}{\sqrt{n-3}},
\] 

Thus, the uncertainty of the correlation estimate depends strongly on the size of the test set. For English, with \(n = 37\), the standard error is \(1/\sqrt{34} \approx 0.17\). If a system obtains an observed Spearman correlation of \(\rho = 0.5\), the Fisher-transformed value gives an approximate 95\% interval which, after transforming back to the correlation scale, is about \([0.21, 0.71]\). This wide interval shows that even a seemingly moderate correlation remains highly imprecise when evaluated on only 37 target words.

These calculations are intended as diagnostic illustrations rather than as exact inferential tests. Nevertheless, they show that small target sets in lexical semantic change benchmarks place strong limits on how confidently one can interpret small differences between systems. This is particularly important when many systems cluster around similar scores in Subtask 1, since the apparent ranking of systems may partly reflect sampling noise rather than robust differences in model quality.

For readers who are less familiar with these statistical concepts, the main point is simple: when a benchmark contains only a small number of target words, evaluation scores are inherently unstable. One or two additional correct or incorrect classifications can noticeably change the accuracy within the test set. Similarly, a seemingly moderate accuracy or rank correlation computed on the \textbf{small test set} may correspond to a wide range of plausible values for the model’s underlying performance on the \textbf{larger population} of possible target words.

\subsubsection{Frequency bias}
Although the organisers have controlled the frequencies of the targets to reduce trivial shortcuts, their own analysis shows that many submitted systems still correlated strongly with frequency variables; for some models, the correlations exceeded 0.8 \citep{schlechtweg_semeval-2020_2020}. 

\subsubsection{Cross-linguistic representativeness}
SemEval 2020 Task 1 was an important early step toward a shared multilingual evaluation of lexical semantic change detection systems. At the same time, because it includes only four European languages (English, German, Latin, and Swedish), it should not be treated as sufficient on its own for evaluating language-independent semantic change detection. The limitation is not that SemEval 2020 failed to solve cross-linguistic representativeness completely, but that results based only on this benchmark may generalise primarily to languages with relatively similar historical resources, writing traditions, and Indo-European structural properties. Methods that perform well on SemEval 2020 may still face additional challenges in typologically different, non-European, lower-resource, or non-Latin-script languages, where tokenisation, morphology, syntactic annotation, diachronic spelling variation, and corpus availability may differ substantially. Therefore, SemEval 2020 is best understood as a foundational benchmark rather than a complete test of cross-linguistic robustness. This limitation has been partly addressed by later datasets and shared tasks, including DIACR-Ita \citep{basile_diacr-ita_2020}, RuShiftEval \citep{kutuzov_rushifteval_2021}, LSCDiscovery \citep{zamora-reina_lscdiscovery_2022}, TempoWiC \citep{loureiro_tempowic_2022}, NorDiaChange \citep{kutuzov_nordiachange_2022}, ChiWUG \citep{chen_chiwug_2023}, and AXOLOTL'24 \citep{fedorova_axolotl24_2024}. Future evaluations should therefore report performance across a broader set of benchmarks with different language families than just Indo-European where possible, rather than relying on SemEval 2020 alone as evidence of general multilingual robustness.

\section{Implications/Applications}
Future datasets and shared tasks for lexical semantic change detection should be designed with much greater theoretical, linguistic, and methodological care.

The first implication concerns how semantic change is operationalised. A possible way forward is not to abandon sense-based benchmarks (as they are still practically useful), but to complement them with more explicitly multidimensional evaluations. One option is to preserve the graded information in usage-relatedness data rather than reducing it too quickly to discrete sense clusters. Another is to evaluate semantic change at multiple descriptive levels: sense inventory change, sense-frequency redistribution, collocational change, constructional change, and discourse-domain change. For example, a benchmark could report whether a model detects not only that a lexeme has changed, but also in what dimensions (e.g., syntactic , collocates, or contextual patterns) and/or what types of change. Such an approach would make lexical semantic change evaluation more compatible with the theoretical view that meaning change is gradual, usage-based, and distributed across recurrent patterns of language use. To our knowledge, no widely adopted lexical semantic change benchmark yet evaluates semantic change simultaneously across sense-inventory, collocational, constructional, and discourse-level dimensions. Several recent resources are moving in this direction, but only partially. AXOLOTL'24 \citep{fedorova_axolotl24_2024} introduces explainable semantic change modeling through novel sense detection and definition generation, and LSC-CTD \citep{cassotti-etal-2024-using} provides labels for causes and types of semantic shift.

Another implication is that future benchmarks should move beyond a strictly two-period design and model semantic change more continuously across time. Restricting evaluation to 2 periods makes annotation manageable, but reduces diachronic development to an endpoint comparison and may miss gradual shifts, intermediate stages, temporary reversals, or changes in the pace of development. More continuous temporal modelling, whether through multiple time slices or trajectory-based evaluation, would allow benchmarks to capture not only how much a lexeme has changed, but also when and how that change unfolds. Several influential modelling approaches already treat semantic change as a trajectory rather than as a simple endpoint comparison. For example, \citet{frermann-lapata-2016-bayesian} model diachronic meaning change as a gradual process in which senses and their prevalence evolve over time, \citet{hamilton-etal-2016-diachronic} use diachronic word embeddings to quantify rates of semantic change across historical time periods. We acknowledge that such designs would substantially increase annotation cost. One possible solution is not to replace expert annotation, but to use large language models as annotation assistants \citep[e.g.][]{Gilardi_2023, ziems-etal-2024-large, tan-etal-2024-large}. For diachronic semantic change, a cautious workflow could therefore begin with LLM-based pilot annotation across multiple time slices, followed by expert verification, correction, and adjudication before scaling up. This human-in-the-loop design would preserve the interpretive reliability required for historical semantic analysis while reducing the cost of constructing richer temporal benchmarks.

The next consideration is that benchmark design also shapes the kinds of models that researchers develop since models are often built to solve the technical challenges posed by available datasets. Future LSC datasets should therefore be designed not only around what is easy to annotate, but also around what kinds of explanatory models the field wants to make possible. Richer benchmarks with temporal, contextual, and interpretable annotations could encourage methods that detect not only whether a word has changed, but also through which linguistic mechanisms the change unfolds.

Another implication concerns data quality and preprocessing. Future datasets should include explicit denoising protocols, preprocessing logs, and released scripts so that users know exactly what was normalised, corrected, or filtered. This is especially important because some systems rely heavily on linguistic annotation, and preprocessing choices can strongly affect downstream results. Reproducibility in this area depends not only on model code, but also on transparent corpus preparation.

Future evaluation should also be more realistic and statistically robust. Small, manually curated target sets make evaluation fragile: one word can change accuracy by several percentage points, and reported rank correlations may have wide uncertainty intervals. Future shared tasks should therefore use larger target inventories, vocabulary-wide discovery settings. In addition, tasks should guard against frequency-based shortcuts, since systems may perform well by tracking corpus frequency rather than semantic change itself. 

Finally, this discussion highlights the need for a broader linguistic coverage. Benchmarks limited to a few European languages cannot support strong claims of language independence. Future datasets should include typologically diverse, lower-resource, and non-Latin-script languages.

\section*{Acknowledgements}
We would like to thank Martijn Windelinckx for his suggestions while working with the Latin dataset.

\section*{Funding Statement}
This project has received funding from the European Union’s Horizon Europe programme for research and innovation under MSCA Doctoral Networks, Grant Agreement No. 101119511.

\section*{Competing interests} 
The authors have no competing interests to declare.

\section*{Data Accessibility Statement}
The SemEval 2020 Task 1 dataset is available at \url{https://www.ims.uni-stuttgart.de/en/research/resources/corpora/sem-eval-ulscd/}. \\ The pre-processing procedure of the SemEval 2020 Task 1 dataset in one of our previous study is available at \url{https://github.com/phantatbach/LChange26-Dep}.
\\ The samples analysed in this study, along with the codes are available at \url{https://doi.org/10.5281/zenodo.19408459}.

\bibliographystyle{johd}
\bibliography{bib}

@inproceedings{hamilton-etal-2016-diachronic,
    title = "Diachronic Word Embeddings Reveal Statistical Laws of Semantic Change",
    author = "Hamilton, William L.  and
      Leskovec, Jure  and
      Jurafsky, Dan",
    editor = "Erk, Katrin  and
      Smith, Noah A.",
    booktitle = "Proceedings of the 54th Annual Meeting of the Association for Computational Linguistics (Volume 1: Long Papers)",
    month = aug,
    year = "2016",
    address = "Berlin, Germany",
    publisher = "Association for Computational Linguistics",
    url = "https://aclanthology.org/P16-1141/",
    doi = "10.18653/v1/P16-1141",
    pages = "1489--1501"
}

@article{frermann-lapata-2016-bayesian,
    title = "A {B}ayesian Model of Diachronic Meaning Change",
    author = "Frermann, Lea  and
      Lapata, Mirella",
    editor = "Lee, Lillian  and
      Johnson, Mark  and
      Toutanova, Kristina",
    journal = "Transactions of the Association for Computational Linguistics",
    volume = "4",
    year = "2016",
    address = "Cambridge, MA",
    publisher = "MIT Press",
    url = "https://aclanthology.org/Q16-1003/",
    pages = "31--45"
}

@inproceedings{tan-etal-2024-large,
    title = "Large Language Models for Data Annotation and Synthesis: A Survey",
    author = "Tan, Zhen  and
      Li, Dawei  and
      Wang, Song  and
      Beigi, Alimohammad  and
      Jiang, Bohan  and
      Bhattacharjee, Amrita  and
      Karami, Mansooreh  and
      Li, Jundong  and
      Cheng, Lu  and
      Liu, Huan",
    editor = "Al-Onaizan, Yaser  and
      Bansal, Mohit  and
      Chen, Yun-Nung",
    booktitle = "Proceedings of the 2024 Conference on Empirical Methods in Natural Language Processing",
    month = nov,
    year = "2024",
    address = "Miami, Florida, USA",
    publisher = "Association for Computational Linguistics",
    url = "https://aclanthology.org/2024.emnlp-main.54/",
    doi = "10.18653/v1/2024.emnlp-main.54",
    pages = "930--957"
}

@article{ziems-etal-2024-large,
    title = "Can Large Language Models Transform Computational Social Science?",
    author = "Ziems, Caleb  and
      Held, William  and
      Shaikh, Omar  and
      Chen, Jiaao  and
      Zhang, Zhehao  and
      Yang, Diyi",
    journal = "Computational Linguistics",
    volume = "50",
    number = "1",
    month = mar,
    year = "2024",
    address = "Cambridge, MA",
    publisher = "MIT Press",
    url = "https://aclanthology.org/2024.cl-1.8/",
    pages = "237--291"
}

@article{Gilardi_2023,
   title={ChatGPT outperforms crowd workers for text-annotation tasks},
   volume={120},
   ISSN={1091-6490},
   url={http://dx.doi.org/10.1073/pnas.2305016120},
   DOI={10.1073/pnas.2305016120},
   number={30},
   journal={Proceedings of the National Academy of Sciences},
   publisher={Proceedings of the National Academy of Sciences},
   author={Gilardi, Fabrizio and Alizadeh, Meysam and Kubli, Maël},
   year={2023},
   month=jul
}

@inproceedings{cassotti-etal-2024-using,
    title = "Using Synchronic Definitions and Semantic Relations to Classify Semantic Change Types",
    author = "Cassotti, Pierluigi  and
      De Pascale, Stefano  and
      Tahmasebi, Nina",
    editor = "Ku, Lun-Wei  and
      Martins, Andre  and
      Srikumar, Vivek",
    booktitle = "Proceedings of the 62nd Annual Meeting of the Association for Computational Linguistics (Volume 1: Long Papers)",
    month = aug,
    year = "2024",
    address = "Bangkok, Thailand",
    publisher = "Association for Computational Linguistics",
    url = "https://aclanthology.org/2024.acl-long.249/",
    doi = "10.18653/v1/2024.acl-long.249",
    pages = "4539--4553"
}

@book{geeraerts_theories_2010,
	address = {Oxford, New York},
	title = {Theories of lexical semantics},
	isbn = {978-0-19-870031-9 978-0-19-870030-2},
	language = {en},
	publisher = {Oxford University Press},
	author = {Geeraerts, Dirk},
	year = {2010},
}

@article{tahmasebi_computational_2021,
	title = {Computational approaches to semantic change},
	language = {en},
	author = {Tahmasebi, Nina and Borin, Lars and Jatowt, Adam and Xu, Yang and Hengchen, Simon},
	year = {2021},
}

@inproceedings{kutuzov_grammatical_2021,
	address = {Online},
	title = {Grammatical {Profiling} for {Semantic} {Change} {Detection}},
	url = {https://aclanthology.org/2021.conll-1.33},
	doi = {10.18653/v1/2021.conll-1.33},
	language = {en},
	urldate = {2025-08-30},
	booktitle = {Proceedings of the 25th {Conference} on {Computational} {Natural} {Language} {Learning}},
	publisher = {Association for Computational Linguistics},
	author = {Kutuzov, Andrey and Pivovarova, Lidia and Giulianelli, Mario},
	year = {2021},
	pages = {423--434},
}

@inproceedings{schlechtweg_semeval-2020_2020,
	address = {Barcelona (online)},
	title = {{SemEval}-2020 {Task} 1: {Unsupervised} {Lexical} {Semantic} {Change} {Detection}},
	shorttitle = {{SemEval}-2020 {Task} 1},
	url = {https://aclanthology.org/2020.semeval-1.1},
	doi = {10.18653/v1/2020.semeval-1.1},
	language = {en},
	urldate = {2025-10-21},
	booktitle = {Proceedings of the {Fourteenth} {Workshop} on {Semantic} {Evaluation}},
	publisher = {International Committee for Computational Linguistics},
	author = {Schlechtweg, Dominik and McGillivray, Barbara and Hengchen, Simon and Dubossarsky, Haim and Tahmasebi, Nina},
	year = {2020},
	pages = {1--23},
}

@incollection{basile_diacr-ita_2020,
	address = {Torino},
	title = {{DIACR}-{Ita} @ {EVALITA2020}: {Overview} of the {EVALITA2020} {Diachronic} {Lexical} {Semantics} ({DIACR}-{Ita}) {Task}},
	copyright = {https://creativecommons.org/licenses/by-nc-nd/4.0/},
	isbn = {979-12-80136-32-9},
	shorttitle = {{DIACR}-{Ita} @ {EVALITA2020}},
	url = {https://books.openedition.org/aaccademia/7613},
	doi = {10.4000/books.aaccademia.7613},
	language = {en},
	urldate = {2025-12-01},
	booktitle = {{EVALITA} {Evaluation} of {NLP} and {Speech} {Tools} for {Italian} - {December} 17th, 2020},
	publisher = {Accademia University Press},
	author = {Basile, Pierpaolo and Caputo, Annalina and Caselli, Tommaso and Cassotti, Pierluigi and Varvara, Rossella},
	editor = {Basile, Valerio and Croce, Danilo and Maro, Maria and Passaro, Lucia C.},
	year = {2020},
	pages = {411--419},
}

@inproceedings{kutuzov_rushifteval_2021,
	title = {{RuShiftEval}: {A} {Shared} {Task} on {Semantic} {Shift} {Detection} for {Russian}},
	doi = {10.28995/2075-7182-2021-20-533-545},
	booktitle = {Computational linguistics and intellectual technologies},
	author = {Kutuzov, Andrey and Pivovarova, Lidia},
	year = {2021},
}

@inproceedings{cassotti_xl-lexeme_2023,
	address = {Toronto, Canada},
	title = {{XL}-{LEXEME}: {WiC} {Pretrained} {Model} for {Cross}-{Lingual} {LEXical} {sEMantic} {changE}},
	url = {https://aclanthology.org/2023.acl-short.135/},
	doi = {10.18653/v1/2023.acl-short.135},
	booktitle = {Proceedings of the 61st {Annual} {Meeting} of the {Association} for {Computational} {Linguistics} ({Volume} 2: {Short} {Papers})},
	publisher = {Association for Computational Linguistics},
	author = {Cassotti, Pierluigi and Siciliani, Lucia and DeGemmis, Marco and Semeraro, Giovanni and Basile, Pierpaolo},
	editor = {Rogers, Anna and Boyd-Graber, Jordan and Okazaki, Naoaki},
	month = jul,
	year = {2023},
	pages = {1577--1585},
}

@inproceedings{schlechtweg_diachronic_2018,
	address = {New Orleans, Louisiana},
	title = {Diachronic {Usage} {Relatedness} ({DURel}): {A} {Framework} for the {Annotation} of {Lexical} {Semantic} {Change}},
	url = {https://aclanthology.org/N18-2027/},
	doi = {10.18653/v1/N18-2027},
	booktitle = {Proceedings of the 2018 {Conference} of the {North} {American} {Chapter} of the {Association} for {Computational} {Linguistics}: {Human} {Language} {Technologies}, {Volume} 2 ({Short} {Papers})},
	publisher = {Association for Computational Linguistics},
	author = {Schlechtweg, Dominik and Schulte im Walde, Sabine and Eckmann, Stefanie},
	editor = {Walker, Marilyn and Ji, Heng and Stent, Amanda},
	month = jun,
	year = {2018},
	pages = {169--174},
}

@book{hanks_lexical_2013,
	title = {Lexical {Analysis}: {Norms} and {Exploitations}},
	isbn = {978-0-262-31285-1},
	url = {https://doi.org/10.7551/mitpress/9780262018579.001.0001},
	doi = {10.7551/mitpress/9780262018579.001.0001},
	publisher = {The MIT Press},
	author = {Hanks, Patrick},
	month = jan,
	year = {2013},
}

@misc{phan-tat_transparent_2026,
	title = {Transparent {Semantic} {Change} {Detection} with {Dependency}-{Based} {Profiles}},
	url = {https://arxiv.org/abs/2601.02891},
	author = {Phan-Tat, Bach and Heylen, Kris and Geeraerts, Dirk and Pascale, Stefano De and Speelman, Dirk},
	year = {2026},
	note = {\_eprint: 2601.02891},
}

@misc{traugott_semantic_2017,
	title = {Semantic {Change}},
	url = {https://oxfordre.com/linguistics/view/10.1093/acrefore/9780199384655.001.0001/acrefore-9780199384655-e-323},
	doi = {10.1093/acrefore/9780199384655.013.323},
	publisher = {Oxford University Press},
	author = {Traugott, Elizabeth Closs},
	month = mar,
	year = {2017},
}

@inproceedings{fedorova_axolotl24_2024,
	address = {Bangkok, Thailand},
	title = {{AXOLOTL}'24 {Shared} {Task} on {Multilingual} {Explainable} {Semantic} {Change} {Modeling}},
	url = {https://aclanthology.org/2024.lchange-1.8/},
	doi = {10.18653/v1/2024.lchange-1.8},
	booktitle = {Proceedings of the 5th {Workshop} on {Computational} {Approaches} to {Historical} {Language} {Change}},
	publisher = {Association for Computational Linguistics},
	author = {Fedorova, Mariia and Mickus, Timothee and Partanen, Niko and Siewert, Janine and Spaziani, Elena and Kutuzov, Andrey},
	editor = {Tahmasebi, Nina and Montariol, Syrielle and Kutuzov, Andrey and Alfter, David and Periti, Francesco and Cassotti, Pierluigi and Huebscher, Netta},
	month = aug,
	year = {2024},
	pages = {72--91},
}

@inproceedings{zamora-reina_lscdiscovery_2022,
	address = {Dublin, Ireland},
	title = {{LSCDiscovery}: {A} shared task on semantic change discovery and detection in {Spanish}},
	url = {https://aclanthology.org/2022.lchange-1.16/},
	doi = {10.18653/v1/2022.lchange-1.16},
	booktitle = {Proceedings of the 3rd {Workshop} on {Computational} {Approaches} to {Historical} {Language} {Change}},
	publisher = {Association for Computational Linguistics},
	author = {Zamora-Reina, Frank D. and Bravo-Marquez, Felipe and Schlechtweg, Dominik},
	editor = {Tahmasebi, Nina and Montariol, Syrielle and Kutuzov, Andrey and Hengchen, Simon and Dubossarsky, Haim and Borin, Lars},
	month = may,
	year = {2022},
	pages = {149--164},
}

@inproceedings{kutuzov_nordiachange_2022,
	address = {Marseille, France},
	title = {{NorDiaChange}: {Diachronic} {Semantic} {Change} {Dataset} for {Norwegian}},
	url = {https://aclanthology.org/2022.lrec-1.274/},
	booktitle = {Proceedings of the {Thirteenth} {Language} {Resources} and {Evaluation} {Conference}},
	publisher = {European Language Resources Association},
	author = {Kutuzov, Andrey and Touileb, Samia and Mæhlum, Petter and Enstad, Tita and Wittemann, Alexandra},
	editor = {Calzolari, Nicoletta and Béchet, Frédéric and Blache, Philippe and Choukri, Khalid and Cieri, Christopher and Declerck, Thierry and Goggi, Sara and Isahara, Hitoshi and Maegaard, Bente and Mariani, Joseph and Mazo, Hélène and Odijk, Jan and Piperidis, Stelios},
	month = jun,
	year = {2022},
	pages = {2563--2572},
}

@misc{loureiro_tempowic_2022,
	title = {{TempoWiC}: {An} {Evaluation} {Benchmark} for {Detecting} {Meaning} {Shift} in {Social} {Media}},
	url = {https://arxiv.org/abs/2209.07216},
	author = {Loureiro, Daniel and D'Souza, Aminette and Muhajab, Areej Nasser and White, Isabella A. and Wong, Gabriel and Anke, Luis Espinosa and Neves, Leonardo and Barbieri, Francesco and Camacho-Collados, Jose},
	year = {2022},
	note = {\_eprint: 2209.07216},
}

@misc{phan-tat_reframe_2026,
	title = {{ReFRAME} or {Remain}: {Unsupervised} {Lexical} {Semantic} {Change} {Detection} with {Frame} {Semantics}},
	url = {https://arxiv.org/abs/2602.04514},
	author = {Phan-Tat, Bach and Heylen, Kris and Geeraerts, Dirk and Pascale, Stefano De and Speelman, Dirk},
	year = {2026},
	note = {\_eprint: 2602.04514},
}

@inproceedings{laicher_explaining_2021,
	address = {Online},
	title = {Explaining and {Improving} {BERT} {Performance} on {Lexical} {Semantic} {Change} {Detection}},
	url = {https://aclanthology.org/2021.eacl-srw.25/},
	doi = {10.18653/v1/2021.eacl-srw.25},
	booktitle = {Proceedings of the 16th {Conference} of the {European} {Chapter} of the {Association} for {Computational} {Linguistics}: {Student} {Research} {Workshop}},
	publisher = {Association for Computational Linguistics},
	author = {Laicher, Severin and Kurtyigit, Sinan and Schlechtweg, Dominik and Kuhn, Jonas and Schulte im Walde, Sabine},
	editor = {Sorodoc, Ionut-Teodor and Sushil, Madhumita and Takmaz, Ece and Agirre, Eneko},
	month = apr,
	year = {2021},
	pages = {192--202},
}

@article{kutuzov_contextualized_2022,
	address = {Linköping, Sweden},
	title = {Contextualized embeddings for semantic change detection: {Lessons} learned},
	volume = {8},
	url = {https://aclanthology.org/2022.nejlt-1.9/},
	doi = {10.3384/nejlt.2000-1533.2022.3478},
	journal = {Northern European Journal of Language Technology},
	publisher = {Linköping University Electronic Press},
	author = {Kutuzov, Andrey and Velldal, Erik and Øvrelid, Lilja},
	editor = {Derczynski, Leon},
	year = {2022},
}

@inproceedings{zhou_finer_2023,
	address = {Tórshavn, Faroe Islands},
	title = {The {Finer} {They} {Get}: {Combining} {Fine}-{Tuned} {Models} {For} {Better} {Semantic} {Change} {Detection}},
	url = {https://aclanthology.org/2023.nodalida-1.52/},
	booktitle = {Proceedings of the 24th {Nordic} {Conference} on {Computational} {Linguistics} ({NoDaLiDa})},
	publisher = {University of Tartu Library},
	author = {Zhou, Wei and Tahmasebi, Nina and Dubossarsky, Haim},
	editor = {Alumäe, Tanel and Fishel, Mark},
	month = may,
	year = {2023},
	pages = {518--528},
}

@inproceedings{ma_graph-based_2024,
	address = {St. Julian's, Malta},
	title = {Graph-based {Clustering} for {Detecting} {Semantic} {Change} {Across} {Time} and {Languages}},
	url = {https://aclanthology.org/2024.eacl-long.93/},
	doi = {10.18653/v1/2024.eacl-long.93},
	booktitle = {Proceedings of the 18th {Conference} of the {European} {Chapter} of the {Association} for {Computational} {Linguistics} ({Volume} 1: {Long} {Papers})},
	publisher = {Association for Computational Linguistics},
	author = {Ma, Xianghe and Strube, Michael and Zhao, Wei},
	editor = {Graham, Yvette and Purver, Matthew},
	month = mar,
	year = {2024},
	pages = {1542--1561},
}

@inproceedings{oda_improving_2025,
	address = {Hanoi, Vietnam},
	title = {Improving {Interpretability} of {Lexical} {Semantic} {Change} with {Neurobiological} {Features}},
	url = {https://aclanthology.org/2025.paclic-1.1/},
	booktitle = {Proceedings of the 39th {Pacific} {Asia} {Conference} on {Language}, {Information} and {Computation}},
	publisher = {Association for Computational Linguistics},
	author = {Oda, Kohei and Takamura, Hiroya and Shirai, Kiyoaki and Kertkeidkachorn, Natthawut},
	editor = {Huang, Chu-Ren and Harada, Yasunari and Kim, Jong-Bok and Huyen, Nguyen T.M. and Huong, Le Thanh and Hien, Pham and Chersoni, Emmanuele and Nguyen, Le Minh and Roxas, Rachel Edita Oñate and Dita, Sherly},
	month = dec,
	year = {2025},
	pages = {1--14},
}

@article{davies_expanding_2012,
	title = {Expanding horizons in historical linguistics with the 400-million word {Corpus} of {Historical} {American} {English}},
	volume = {7},
	url = {https://doi.org/10.3366/cor.2012.0024},
	doi = {10.3366/cor.2012.0024},
	number = {2},
	journal = {Corpora},
	author = {Davies, Mark},
	year = {2012},
	note = {\_eprint: https://doi.org/10.3366/cor.2012.0024},
	pages = {121--157},
}

@inproceedings{alatrash_ccoha_2020,
	address = {Marseille, France},
	title = {{CCOHA}: {Clean} {Corpus} of {Historical} {American} {English}},
	isbn = {979-10-95546-34-4},
	url = {https://aclanthology.org/2020.lrec-1.859/},
	language = {eng},
	booktitle = {Proceedings of the {Twelfth} {Language} {Resources} and {Evaluation} {Conference}},
	publisher = {European Language Resources Association},
	author = {Alatrash, Reem and Schlechtweg, Dominik and Kuhn, Jonas and Schulte im Walde, Sabine},
	editor = {Calzolari, Nicoletta and Béchet, Frédéric and Blache, Philippe and Choukri, Khalid and Cieri, Christopher and Declerck, Thierry and Goggi, Sara and Isahara, Hitoshi and Maegaard, Bente and Mariani, Joseph and Mazo, Hélène and Moreno, Asuncion and Odijk, Jan and Piperidis, Stelios},
	month = may,
	year = {2020},
	pages = {6958--6966},
}

@misc{deutsches_textarchiv_grundlage_2017,
	address = {Berlin},
	title = {Grundlage für ein {Referenzkorpus} der neuhochdeutschen {Sprache}},
	url = {http://www.deutschestextarchiv.de/},
	author = {{Deutsches Textarchiv}},
	editor = {{Berlin-Brandenburgische Akademie der Wissenschaften}},
	year = {2017},
}

@misc{staatsbibliothek_zu_berlin_berliner_2018,
	title = {Berliner {Zeitung} {Diachronic} {Newspaper} {Corpus}},
	publisher = {Staatsbibliothek zu Berlin – Preußischer Kulturbesitz},
	author = {{Staatsbibliothek zu Berlin}},
	year = {2018},
}

@misc{staatsbibliothek_zu_berlin_neues_2018,
	title = {Neues {Deutschland} {Diachronic} {Newspaper} {Corpus}},
	publisher = {Staatsbibliothek zu Berlin – Preußischer Kulturbesitz},
	author = {{Staatsbibliothek zu Berlin}},
	year = {2018},
}

@incollection{mcgillivray_tools_2013,
	address = {Tübingen},
	series = {Korpuslinguistik und interdisziplinäre {Perspektiven} auf {Sprache}},
	title = {Tools for historical corpus research, and a corpus of {Latin}},
	volume = {3},
	booktitle = {New {Methods} in {Historical} {Corpus} {Linguistics}},
	publisher = {Narr},
	author = {McGillivray, Barbara and Kilgarriff, Adam},
	editor = {Bennett, Paul and Durrell, Martin and Scheible, Silke and Whitt, Richard J.},
	year = {2013},
	pages = {247--257},
}

@misc{sprakbanken_text_kubhist_2019,
	title = {The {Kubhist} {Corpus}, v2},
	url = {https://spraakbanken.gu.se/en/resources/kubhist2},
	publisher = {Department of Swedish, University of Gothenburg},
	author = {{Språkbanken Text}},
	year = {2019},
	note = {Published: Data set},
}

@misc{kilgarriff_i_1997,
	title = {"{I} don't believe in word senses"},
	url = {https://arxiv.org/abs/cmp-lg/9712006},
	author = {Kilgarriff, Adam},
	year = {1997},
	note = {\_eprint: cmp-lg/9712006},
}

@book{cruse_lexical_1986,
	address = {Cambridge},
	title = {Lexical {Semantics}},
	isbn = {978-0-521-27643-6},
	publisher = {Cambridge University Press},
	author = {Cruse, D. A.},
	year = {1986},
}

@misc{sprakbanken_text_semeval2020_2024,
	title = {{SemEval2020} {Task} 1},
	url = {https://spraakbanken.gu.se/en/resources/semeval2020},
	doi = {10.23695/d79w-qa67},
	publisher = {Språkbanken Text},
	author = {{Språkbanken Text}},
	year = {2024},
}

@misc{mcgillivray_latinise_2020,
	title = {{LatinISE} test data for {SemEval} 2020 task 1 with additional token versions of the corpora},
	copyright = {Creative Commons Attribution 4.0 International, Open Access},
	url = {https://zenodo.org/record/3992738},
	doi = {10.5281/ZENODO.3992738},
	language = {la},
	urldate = {2026-03-17},
	publisher = {Zenodo},
	author = {McGillivray, Barbara and Schlechtweg, Dominik and Dubossarsky, Haim and Tahmasebi, Nina and Hengchen, Simon},
	month = aug,
	year = {2020},
}

@inproceedings{chen_chiwug_2023,
	address = {Singapore},
	title = {{ChiWUG}: {A} {Graph}-based {Evaluation} {Dataset} for {Chinese} {Lexical} {Semantic} {Change} {Detection}},
	url = {https://aclanthology.org/2023.lchange-1.10/},
	doi = {10.18653/v1/2023.lchange-1.10},
	booktitle = {Proceedings of the 4th {Workshop} on {Computational} {Approaches} to {Historical} {Language} {Change}},
	publisher = {Association for Computational Linguistics},
	author = {Chen, Jing and Chersoni, Emmanuele and Schlechtweg, Dominik and Prokic, Jelena and Huang, Chu-Ren},
	editor = {Tahmasebi, Nina and Montariol, Syrielle and Dubossarsky, Haim and Kutuzov, Andrey and Hengchen, Simon and Alfter, David and Periti, Francesco and Cassotti, Pierluigi},
	month = dec,
	year = {2023},
	pages = {93--99},
}

@inproceedings{umarova_current_2025,
	address = {Suzhou, China},
	title = {Current {Semantic}-change {Quantification} {Methods} {Struggle} with {Discovery} in the {Wild}},
	isbn = {979-8-89176-332-6},
	url = {https://aclanthology.org/2025.emnlp-main.1791/},
	doi = {10.18653/v1/2025.emnlp-main.1791},
	booktitle = {Proceedings of the 2025 {Conference} on {Empirical} {Methods} in {Natural} {Language} {Processing}},
	publisher = {Association for Computational Linguistics},
	author = {Umarova, Khonzoda and Lee, Lillian and Kim, Laerdon},
	editor = {Christodoulopoulos, Christos and Chakraborty, Tanmoy and Rose, Carolyn and Peng, Violet},
	month = nov,
	year = {2025},
	pages = {35354--35367},
}

@incollection{geeraerts_semantic_2020,
	title = {Semantic {Change}},
	isbn = {978-1-118-78851-6},
	url = {https://onlinelibrary.wiley.com/doi/abs/10.1002/9781118788516.sem042},
	doi = {https://doi.org/10.1002/9781118788516.sem042},
	booktitle = {The {Wiley} {Blackwell} {Companion} to {Semantics}},
	publisher = {John Wiley \& Sons, Ltd},
	author = {Geeraerts, Dirk},
	year = {2020},
	note = {\_eprint: https://onlinelibrary.wiley.com/doi/pdf/10.1002/9781118788516.sem042},
	pages = {1--24},
}

@article{geeraerts_vaguenesss_1993,
	title = {Vagueness's puzzles, polysemy's vagaries},
	volume = {4},
	url = {https://doi.org/10.1515/cogl.1993.4.3.223},
	doi = {doi:10.1515/cogl.1993.4.3.223},
	number = {3},
	urldate = {2026-04-02},
	journal = {Cognitive Linguistics},
	author = {Geeraerts, Dirk},
	year = {1993},
	pages = {223--272},
}

\section*{Supplementary Files (optional)}
\begin{itemize}
    \item Code, data and results for the quality check of the data: \url{https://doi.org/10.5281/zenodo.19408459}
    \item The code and correction procedure used to reproduce our pre-processing steps in our previous study: \url{https://github.com/phantatbach/LChange26-Dep}
\end{itemize}

\end{document}